\documentclass[colorlinks,upint,subscriptcorrection,varvw,hyphenate]{asmeconf}

\hypersetup{%
  pdfauthor={Nehal Afifi, Klaus Rappenecker, Florian Peinhardt, Gerhard Kachel, Alfred Isele, Sven Matthiesen},
  pdftitle={PreForceVision: Physics-Guided Optical Estimation of Bolt Preload from Bolt-Head Deformation},
  pdfkeywords={bolt preload, computer vision, digital image correlation, deformation measurement, bolted joints},
  pdfsubject={Physics-guided optical framework for bolt preload estimation}
}

\allowdisplaybreaks
\usepackage{amsmath}
\usepackage{graphicx}
\usepackage{booktabs}
\usepackage{siunitx}
\usepackage{url}

\begin{document}

\ConfName{Proceedings of the ASME 2026\linebreak International Mechanical Engineering Congress and Exposition}
\ConfAcronym{IMECE2026}
\ConfDate{November 8--12, 2026}
\ConfCity{Vancouver, Canada}
\PaperNo{IMECE2026-192041}

\title{PreForceVision: Physics-Guided Optical Estimation of Bolt Preload from Bolt-Head Deformation}

\SetAuthors{%
  Nehal Afifi\affil{1}\CorrespondingAuthor{nehal.afifi@kit.edu},
  Klaus Rappenecker\affil{1}\affil{2},
  Florian Peinhardt\affil{2},
  Gerhard Kachel\affil{2},
  Alfred Isele\affil{2},
  Sven Matthiesen\affil{1}\CorrespondingAuthor{sven.matthiesen@kit.edu}
}

\SetAffiliation{1}{IPEK - Institute of Product Engineering at the Karlsruhe Institute of Technology (KIT), Karlsruhe, Germany}
\SetAffiliation{2}{Offenburg University of Applied Sciences, Department of Mechanical and Process Engineering, Offenburg, Germany}

\maketitle


\begin{abstract} Bolt preload is a decisive functional quantity in bolted joints because it determines whether external loads are transmitted primarily through frictional clamping or directly through the bolt. Existing preload measurement techniques, including strain-gauge bolts, force washers, ultrasonic methods, and bolt-length measurements, are effective in certain applications but require additional hardware, coupling, calibration, or joint modifications. This limits their scalability for large numbers of standard fasteners and motivates a non-contact optical approach. This paper proposes PreForceVision, a physics-guided optical framework for estimating bolt preload from preload-induced surface deformation of the bolt-head front face. The method treats preload estimation as an inverse deformation-to-force problem, where deformation of the optically accessible bolt-head surface is extracted from reference-loaded image pairs and related to the internal preload state. Preliminary results for an M16 strength class 8.8 bolt show an approximately \SI{12}{\micro\meter} simulated head-face depression at \SI{90}{\kilo\newton}, strain-gauge measurements of comparable magnitude, and an optical 3D measured central depression of approximately \SIrange{10}{20}{\micro\meter}. The contribution is a physics-guided optical methodology combining controlled image-pair acquisition, DIC- or optical-flow-based deformation extraction, multimodal validation, and uncertainty-aware regression for rough preload estimation under controlled conditions.
\end{abstract}

\begin{nomenclature}
\EntryHeading{Roman letters}
\entry{$F_p$}{bolt preload force}
\entry{$\hat{F}_p$}{estimated bolt preload force}
\entry{$I_0, I_L$}{reference and loaded images}
\entry{$M$}{bolt-head mask}
\entry{$S$}{preload-sensitive deformation signature}
\entry{$u,v$}{in-plane displacement components}
\entry{$w$}{out-of-plane displacement or height difference}
\entry{$Q$}{image or correlation quality indicator}
\entry{$\mathbf{x}_m$}{known metadata vector}
\EntryHeading{Greek letters}
\entry{$\varepsilon_r$}{radial strain}
\entry{$\varepsilon_\theta$}{circumferential strain}
\entry{$\sigma_{\hat{F}}$}{uncertainty of preload estimate}
\entry{$\Gamma$}{deformation symmetry descriptor}
\end{nomenclature}

\section{Introduction}
Bolts, in particular, are among the most commonly used machine elements for component connection \cite{ste12} and are used wherever a detachable connection must reliably transmit load. Their mechanical function depends primarily on the preload generated during assembly. If the preload is insufficient, the clamped parts can separate earlier, the fraction of external load carried by the bolt increases, and the fatigue-critical stress amplitude in the bolt rises. Reliable preload assessment is therefore central to the design and maintenance of safety-relevant bolted joints. 
The nominal preload and load-bearing capacity of highly loaded bolted joints are commonly verified in accordance with VDI 2230 \cite{vdi2230}. The most critical joints typically bear the heaviest loads. The actual preload force, crucial for the bolted connection’s functionality, is achieved by applying the tightening torque during assembly. However, the tightening factor varies significantly across assembly methods, as determined by the assembly tool and the control measurement variable \cite{wie07}.

In practice, preload is often controlled indirectly by tightening torque. This is convenient but not a direct measure of bolt force, because a large portion of the applied torque is consumed by thread and bearing-surface friction\cite{vdi2230}. Variations in lubrication, coatings, surface roughness, repeated tightening, and assembly tool behavior can therefore lead to significant preload scatter. As a result, the actual preload state of an assembled joint may remain uncertain, even though it is the decisive functional quantity governing load transfer and fatigue behavior. Once the joint is assembled, the challenge becomes more severe: the preload is not directly visible, and direct measurement typically requires specialized hardware. This motivates inspection concepts that can estimate preload-related information from accessible, non-instrumented surfaces.

Established preload-measurement approaches include sensorized bolts, force-measuring washers, strain gauges, load cells, fiber-optic sensors, ultrasonic techniques, and bolt-length measurements. These methods are important for calibration, laboratory validation, and selected high-value applications, but they typically require additional hardware, wiring, coupling, calibration, or modification of the fastener or joint. For large numbers of standard fasteners, this limits scalability.

Computer vision has been investigated for bolt inspection, especially for detecting missing bolts, corrosion, visible looseness, or rotation of marked bolts \cite{afifi2024data}. These methods are useful for inspection, but most of them estimate a visible state such as bolt rotation or loosening angle rather than the mechanical preload itself \cite{cha2016vision,pham2020bolt,yu2021detection,sun2022yolo,qian2024loosening,malik2024review}. A bolt can lose a functionally relevant amount of preload before a large visible rotation or separation gap is detectable. This motivates a different visual target: not the rotation of the bolt, but the preload-induced surface deformation of the visible bolt-head face.

This paper proposes \emph{PreForceVision}, a physics-guided optical framework for estimating bolt preload from the deformation of the bolt-head front face. The central hypothesis is that axial preload produces a repeatable, micrometer-scale deformation and strain field on the bolt-head surface. If this field can be measured or approximated from reference-loaded image pairs, it can serve as a preload-sensitive optical signature. The proposed approach is a feasibility framework. It first asks whether the mechanical signal is observable, then how it can be extracted from images, and how it can be mapped to preload under controlled conditions.


\begin{figure*}[t]
   \centering
    \includegraphics[width=0.95\textwidth, keepaspectratio, trim=0.cm 0cm 0cm 0cm, clip]{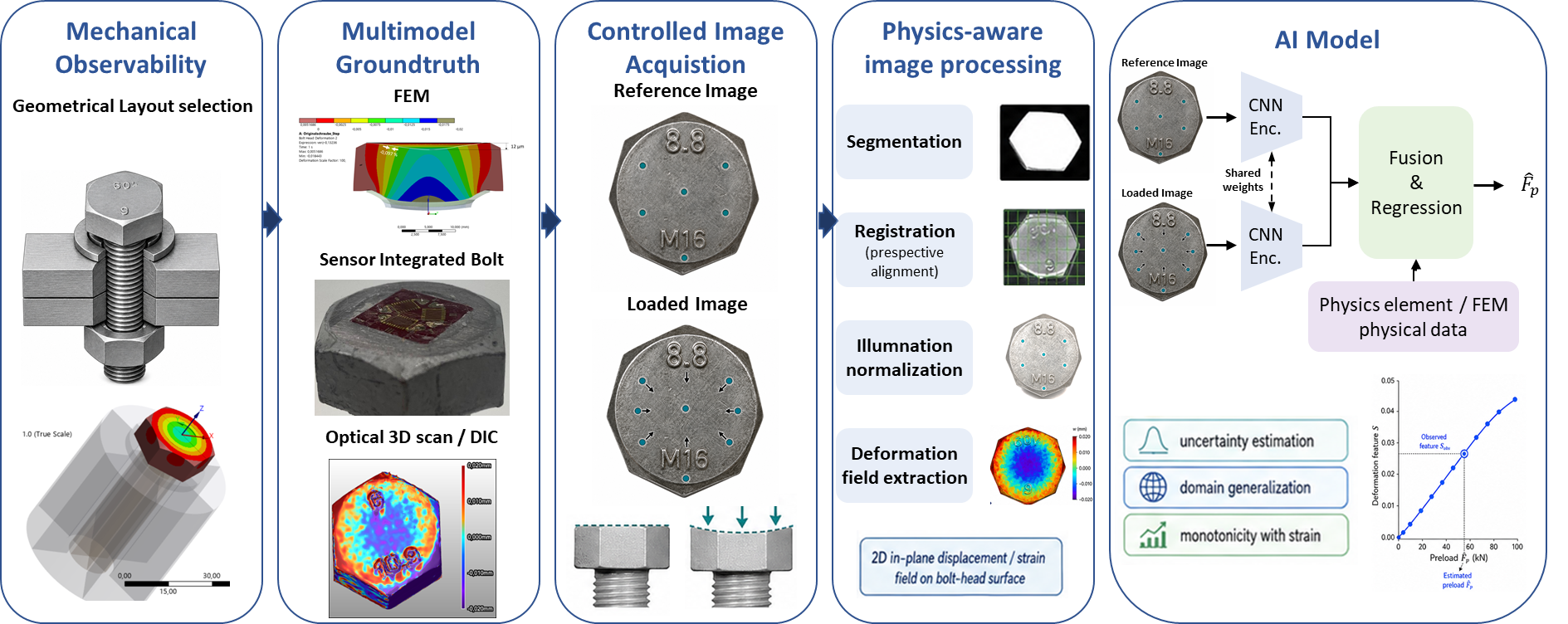}
    \caption{PreForceVision Framework: The proposed framework compares a reference and a loaded image of the bolt-head front face, extracts a preload-sensitive deformation signature, and maps this signature to preload with uncertainty. Hidden assembly geometry is not inferred directly from the head image.}
    \label{fig:concept}
\end{figure*}

\section{Related Work}

\subsection{Preload Monitoring in Bolted Joints}

Direct preload monitoring has been addressed using contact sensors and wave-based approaches. Strain-gauge bolts and load cells provide direct force-related measurements, while ultrasonic and coda-wave techniques use wave propagation changes to infer preload or preload loss \cite{chen2022coda,chen2023fullrange}. Fiber Bragg grating sensors and other embedded sensing concepts have also been proposed for connection-state monitoring \cite{deng2022fbg}. These approaches can be accurate, but they require additional sensing hardware, signal-conditioning infrastructure, and calibration. The present study does not aim to replace these methods in high-accuracy applications; rather, it investigates whether an optical approach can provide a scalable rough estimate of preload for standard fasteners under controlled conditions.

\subsection{Bolt-Head Strain as a Preload Indicator}

The mechanical basis of the proposed framework is supported by previous work showing that the bolt-head surface contains preload-sensitive strain information. Wang et al. investigated the strain distribution on the bolt-head top surface and showed that radial strain has an approximately linear relationship with pretightening force \cite{wang2020bolt}. Their results indicate that the bolt-head face is not only a convenient mounting location for a strain gauge, but also a mechanically informative surface. Wang et al. later proposed a wireless bolt-head strain sensor with self-temperature compensation, using the different sensitivities of radial and circumferential strain regions on the bolt-head face \cite{wang2024head}. Ye et al. also reported a bolt-force monitoring method based on radial strain on the bolt-head surface in maglev infrastructure maintenance, with strain gauges arranged on a circular ring to reduce the influence of lateral forces and bending moments \cite{ye2024maglev}. These studies provide an important premise for PreForceVision: preload produces a spatially structured response on the visible bolt-head face.

\subsection{Digital Image Correlation and Optical Deformation Measurement}

Digital image correlation (DIC) is a full-field optical method for estimating displacement and strain from reference and deformed images. Pan et al. provide a comprehensive review of two-dimensional DIC, including subset matching, displacement estimation, strain computation, and accuracy considerations \cite{pan2009dic}. DIC has also been used in high-accuracy optical extensometry and mechanical testing \cite{zhu2017extensometer}. In the bolt context, Huo et al. proposed a DIC-based bolt preload monitoring method in which an industrial camera records speckle images of the bolt head before and after deformation, and the resulting surface strain field is related to preload \cite{huo2026dic}. Li et al. similarly proposed early bolt-loosening detection using DIC on a bolt-head surface to predict axial force from average strain \cite{li2024dic}. These studies are close to the present work, but they rely on controlled conditions and intentionally textured surfaces. PreForceVision uses DIC not only as a measurement method, but also as a physically interpretable baseline and supervision source for a later image-based predictive model.

\subsection{Vision-Based Bolt Inspection and Deep Learning}

Vision-based bolt monitoring has largely focused on visible loosening, rotation, or damage. Early work combined Hough-transform features and support vector machines for loosened-bolt detection \cite{cha2016vision}. Later studies used deep detectors and synthetic data to identify bolts and bolt-loosening states \cite{pham2020bolt,ta2020monitoring,ta2022corroded,sun2022yolo, afifi2026gaussian}. Other work used keypoint detection, angle estimation, or geometric imaging theory to quantify loosening angles or exposed-bolt lengths \cite{gong2022quantitative,qian2024loosening}. These approaches demonstrate the feasibility of image-based inspection under varying distances, angles, and lighting, but their target is generally a visible geometric state, not preload. Recent review literature confirms that vision-based deep learning for bolted joints is dominated by damage, looseness, and detection tasks rather than quantitative preload estimation \cite{malik2024review}.

Deep learning has also been applied to optical deformation measurement\cite{afifi2025data}. Boukhtache et al. introduced a learning-based approach that connects deep networks with DIC-like displacement estimation \cite{boukhtache2021strainnet}. Yang et al. proposed Deep DIC for end-to-end displacement and strain estimation \cite{yang2022deepdic}. These studies suggest that learned models can accelerate or approximate deformation-field estimation, but they do not remove the need for careful physics-based validation. In the present application, the signal is small, the surface is metallic, and lighting or registration errors can easily dominate the apparent difference between two images. The learning problem must therefore be constrained by mechanics and evaluated against ground truth from finite element analysis, strain gauges, optical 3D measurements, and DIC.

\subsection{Research Gap and Question}
The research gap lies in the missing link between surface-accessible optical measurement and mechanically meaningful internal state estimation. Sensor-based preload methods can provide accurate force-related measurements, but they require modified fasteners, additional hardware, wiring, coupling, or calibration. Vision-based bolt inspection is scalable, but it mainly detects visible states such as missing bolts, corrosion, rotation, or looseness, rather than preload as the functional quantity of the joint. DIC-based methods provide a stronger physical basis, but they still require controlled surface preparation, stable acquisition, and independent validation.

This paper therefore investigates whether the preload-induced deformation of the visible bolt-head front face can be used as an optical signature for rough preload estimation. The research question is: How can small preload-induced surface deformations be extracted and validated systematically so that they support controlled, uncertainty-aware preload estimation?


\section{Problem Formulation}
The problem addressed in this work is to determine whether the preload force of an assembled bolt can be estimated from the small deformation observed on the visible front face of the bolt head after loading. The proposed solution, therefore, does not treat the task as ordinary bolt-image classification. Instead, it compares two images of the same bolt head: a reference image before loading, or at a known baseline state, and a subsequent image after preload application or at a later monitoring state. The objective is to extract the local deformation signature between these two states and relate it to the corresponding preload force.

Accordingly, the input is formulated as an image pair,
\begin{equation}
    \mathcal{I}=\left(I_0,I_L\right)
    \label{eq:image-pair input}
\end{equation}
where $I_0$ denotes the reference image and $I_L$ denotes the loaded or monitored image. The desired output is
\begin{equation}
    \left(\hat{F}_p,\sigma_{\hat{F}}\right)=f_\theta\left(I_0,I_L,\mathbf{x}_m\right)
    \label{eq:mapping}
\end{equation}
where $\hat{F}_p$ is the estimated preload force, $\sigma_{\hat{F}}$ is the associated prediction uncertainty, and $\mathbf{x}_m$ contains known metadata such as bolt type, surface preparation, imaging configuration, and acquisition settings. In the preliminary study, the joint geometry is kept fixed. If multiple bolt or washer geometries are investigated in future work, these parameters are supplied as known metadata rather than inferred from the bolt-head image.

The optical signal used for estimation is the local motion of surface texture, speckle patterns, or identifiable image features on the front face of the bolt head. After rigid registration of the reference and loaded images, the remaining local motion is interpreted as a deformation-related displacement field,
\begin{equation}
    \mathbf{u}(x,y)=\left[u(x,y),v(x,y)\right]^T
    \label{eq:displacment}
\end{equation}
which can be estimated using digital image correlation, optical flow, or learned correspondence. Strain-like quantities are then obtained from displacement gradients, for example
\begin{equation}
    \varepsilon_{xx}=\frac{\partial u}{\partial x},\qquad
    \varepsilon_{yy}=\frac{\partial v}{\partial y},\qquad
    \gamma_{xy}=\frac{\partial u}{\partial y}+\frac{\partial v}{\partial x}
    \label{eq:strain from displacement gradients}
\end{equation}

Because the bolt-head front face has an approximately circular measurement region inside the hexagonal head boundary, the deformation descriptors are also expressed relative to the head center. Relevant descriptors include radial displacement $u_r$, radial strain $\varepsilon_r$, annular strain contrast, center-to-edge deformation contrast, and deformation symmetry. These descriptors form a compact preload-sensitive signature,
\begin{equation}
    S = \Phi\left(\mathbf{u},\nabla \mathbf{u},M,Q\right)
    \label{eq:signature}
\end{equation}
where $M$ is the segmented bolt-head mask and $Q$ contains image-quality, registration-quality, and correlation-quality indicators. The final estimation task is therefore to learn a mapping from this deformation signature, together with known metadata, to preload force and uncertainty.

\subsection{Contribution}
This paper introduces an image-based preload estimation framework named PreForceVision. The novelty lies in treating preload estimation as a mechanically constrained inverse problem, where deformation of the optically accessible bolt-head surface is linked to the internal preload state.

The first contribution is a deformation-to-force formulation based on the preload-induced deformation of the bolt head's front face. The second contribution is a multimodal validation strategy combining finite-element analysis, strain-gauge measurements, optical 3D measurements, and DIC- or optical-flow-based deformation extraction. The third contribution is an uncertainty-aware regression framework that maps physically interpretable deformation descriptors to rough preload estimates under controlled acquisition conditions. The aim is a structured feasibility route toward scalable preload assessment in applications where sensorizing every bolted joint is impractical.

\section{Preliminary Investigation}
The framework is based on the following working hypothesis: Axial preload produces a repeatable deformation pattern on the bolt-head front face; if this pattern can be detected with sufficient signal-to-noise ratio, it can serve as an optical indicator for preload estimation. Therefore, a preliminary analysis was conducted to test whether the expected deformation magnitude is physically observable.
This sequence is necessary because the optical learning task is only meaningful if the underlying mechanical signal can first be confirmed independently. The preliminary investigation therefore does not aim to demonstrate a complete predictive model. Instead, it evaluates whether preload produces a repeatable and measurable deformation of the bolt-head front face that can later serve as a physically interpretable input for image-based estimation.

\subsection{Geometric layout Joint and Measurement Region}
The initial reference case is an M16 ISO 4017 hex-head bolt of strength class 8.8. The selected optical measurement region is the visible bolt-head front face, because it remains accessible after assembly and contains the surface on which preload-induced strain and deformation can be observed. The washer, clamped plates, nut, and shank are not treated as visually measured regions in the first study. Their dimensions are controlled by the test setup and, if required, can be supplied as metadata.
\begin{figure}[t]
    \centering
    \includegraphics[width=0.95\linewidth, trim=0cm 0cm 0.5cm 0cm, clip, keepaspectratio]{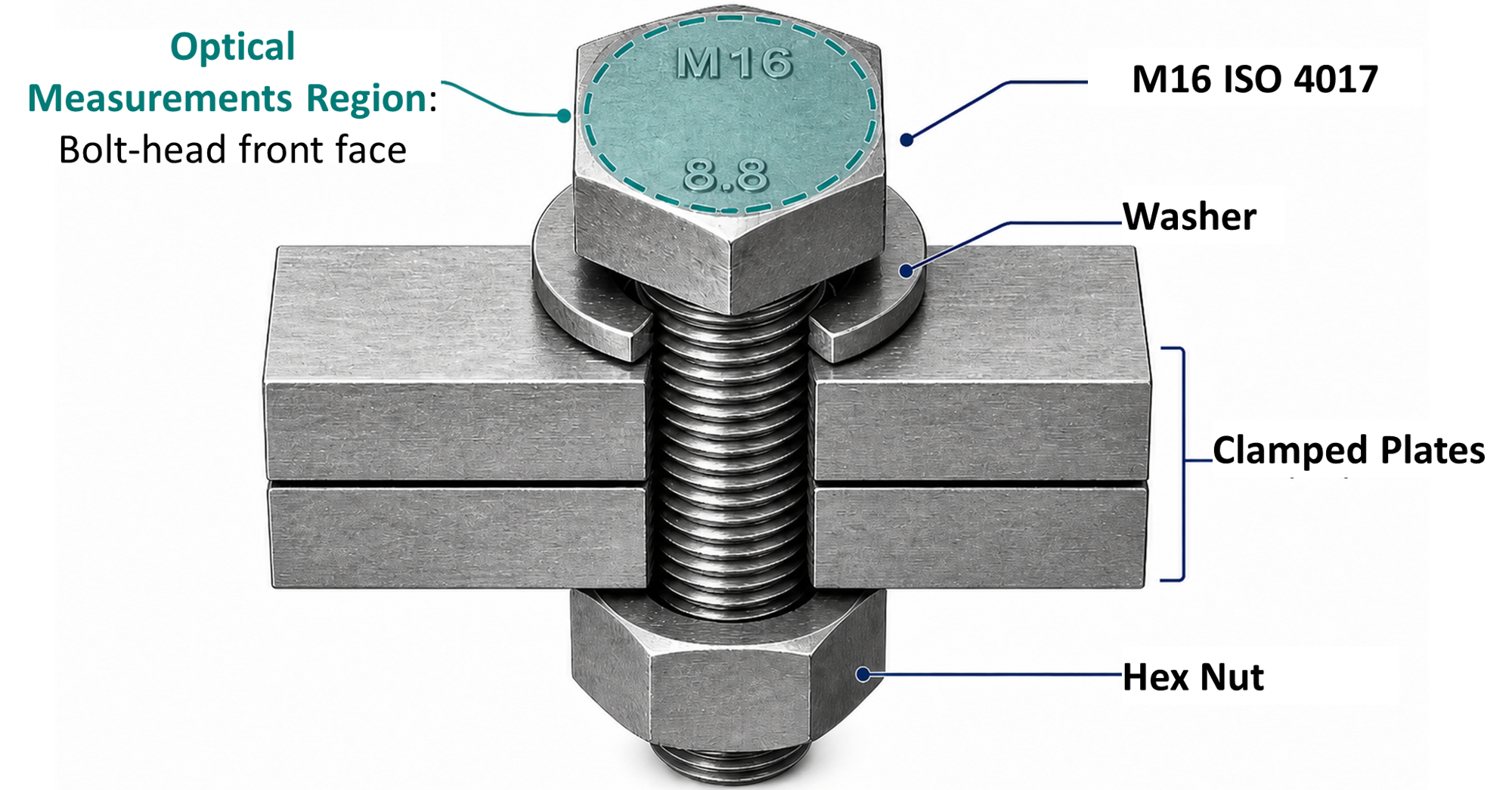}
    \caption{Bolted-joint geometrical layout and optical measurement region. The study focuses on the deformation of the visible front face of the bolt head while the joint geometry is kept fixed.}
    \label{fig:layout}
\end{figure}
\subsection{Mechanical Observability}
\begin{figure*}[t]
\centering
\begin{minipage}{0.48\textwidth}
\centering
\includegraphics[width=\linewidth]{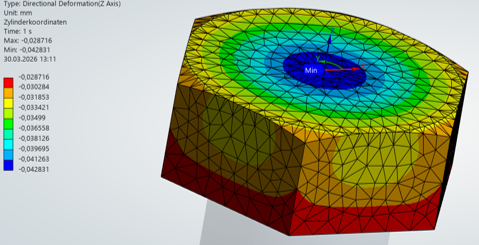}
\end{minipage}\hfill
\begin{minipage}{0.48\textwidth}
\centering
\includegraphics[width=0.92\linewidth]{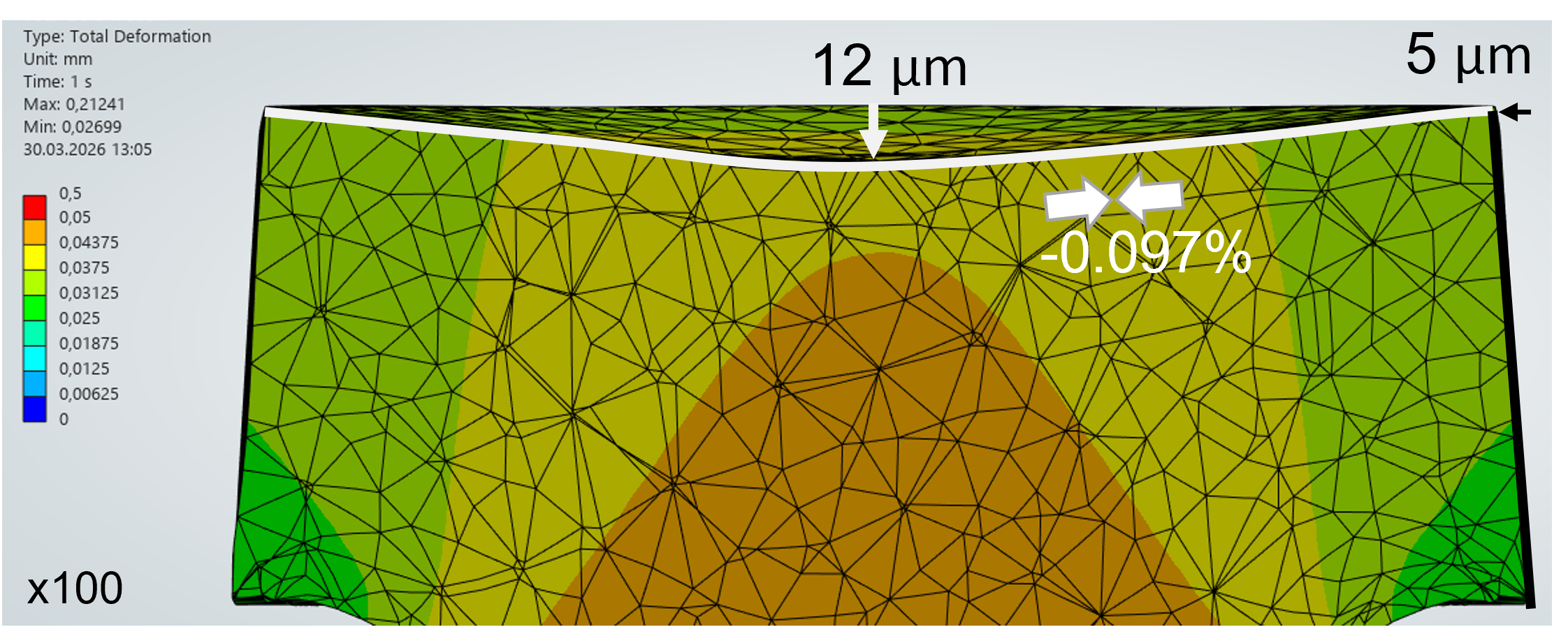}
\end{minipage}
\caption{Preliminary finite-element evidence of preload-induced bolt-head deformation for an M16 property class 8.8 bolt at \SI{90}{\kilo\newton}. The results indicate an approximately \SI{12}{\micro\meter} depression of the bolt-head front face and approximately \SI{0.1}{\percent} surface deformation. The plotted deformation is used here to illustrate the spatial pattern and is not a field-ready optical measurement.}
\label{fig:prelim_fem}
\end{figure*}

The feasibility of PreForceVision depends first on the physical observability of the bolt-head deformation. Preliminary results were obtained on the given geometrical layout. First, a finite-element simulation applied a preload force of \SI{90}{\kilo\newton}, corresponding approximately to \SI{90}{\percent} of the yield strength for the selected strength class. The simulation indicates a concave deformation of the bolt-head front face, with an approximate central depression of \SI{12}{\micro\meter}. The corresponding surface deformation is approximately \SI{0.1}{\percent}. Figure~\ref{fig:prelim_fem} shows the simulated total and directional deformation patterns. The deformation was visually amplified for interpretation; the physical displacement magnitude remains in the micrometer range.

\begin{figure}
    \centering
    \includegraphics[width=1.0\linewidth, keepaspectratio, trim= 7.6cm 4.7cm 7.6cm 4.7cm, clip]{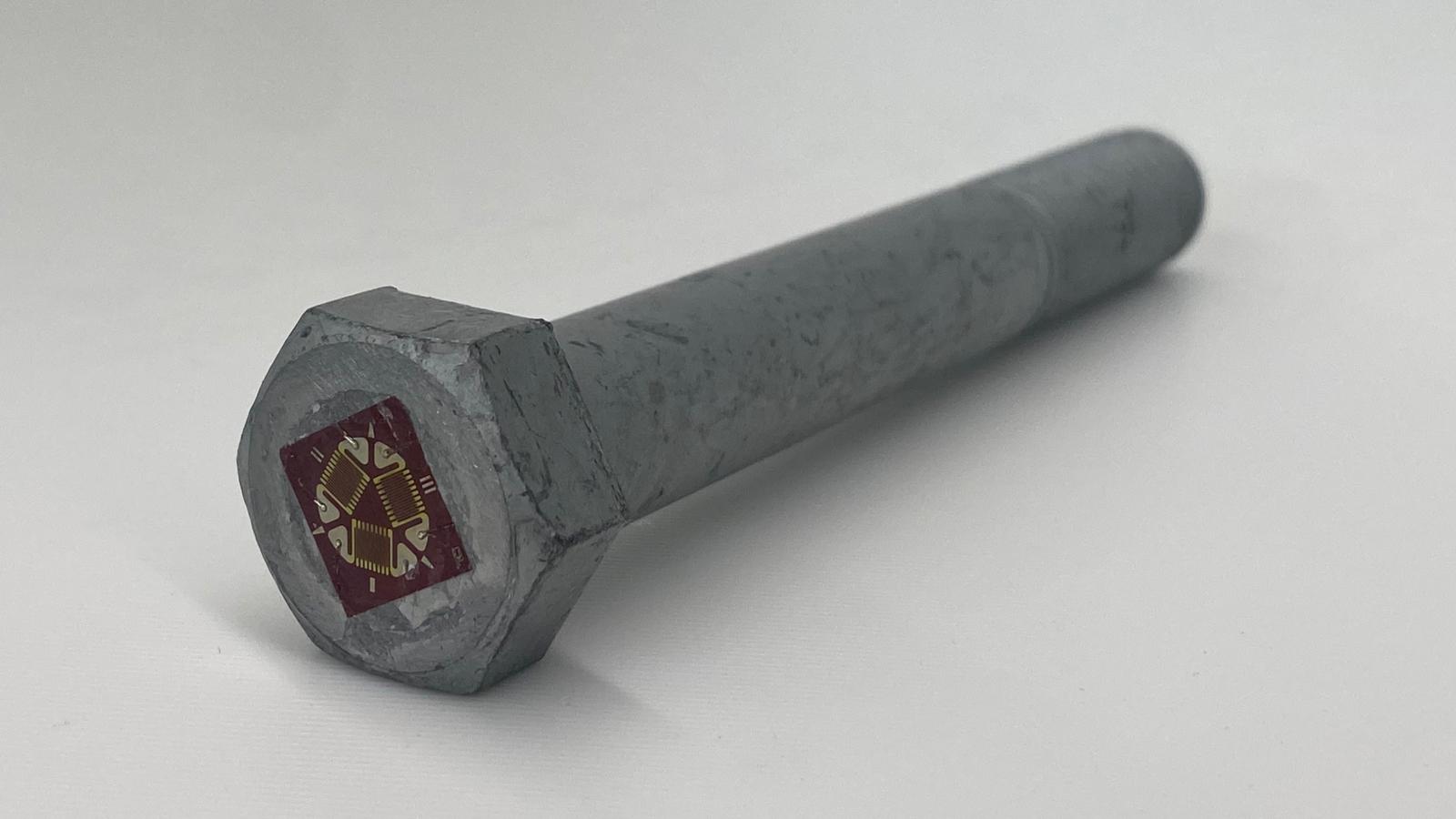}
    \caption{Sensor-Integrated Bolt used in preliminary investigation: strain-gauge measurements on the bolt head}
    \label{fig:sensor-integrated}
\end{figure}

These simulated compressions are consistent with strain-gauge measurements, which were further obtained as a second preliminary analysis from a sensor-integrated bolt. Three strain gauges mounted on the bolt head, seen in Figure~\ref{fig:sensor-integrated}, were evaluated at preload levels of \SI{30}{\kilo\newton}, \SI{60}{\kilo\newton}, and \SI{90}{\kilo\newton}. The mean measured strain increased in magnitude from \SI{-0.046}{\percent} at \SI{30}{\kilo\newton} to \SI{-0.116}{\percent} at \SI{90}{\kilo\newton}, as summarized in Table~\ref{tab:strain_gauge}. The agreement in order of magnitude between simulation and strain measurement supports the assumption that the bolt-head face contains a preload-sensitive mechanical signature.
\begin{table}[t]
\caption{Preliminary strain-gauge measurements on the bolt head. Negative values indicate compressive strain on the measured surface.}
\label{tab:strain_gauge}
\centering
\begin{tabular}{ccccc}
\toprule
Preload & Gauge 1 & Gauge 2 & Gauge 3 & Mean \\
\si{\kilo\newton} & \si{\percent} & \si{\percent} & \si{\percent} & \si{\percent} \\
\midrule
30 & -0.056 & -0.035 & -0.048 & -0.046 \\
60 & -0.084 & -0.069 & -0.094 & -0.082 \\
90 & -0.129 & -0.102 & -0.117 & -0.116 \\
\bottomrule
\end{tabular}
\end{table}

\begin{figure*}[t]
\centering
\begin{minipage}{0.45\textwidth}
\centering
\includegraphics[width=0.6\linewidth]{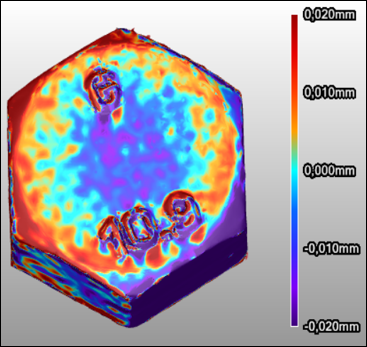}
\end{minipage}\hfill
\begin{minipage}{0.45\textwidth}
\centering
\includegraphics[width=0.6\linewidth]{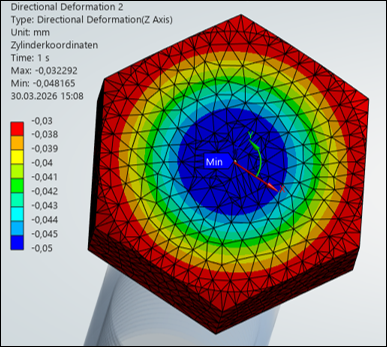}\hfill
\end{minipage}
\caption{Preliminary optical 3D comparison of the bolt-head front face before and after assembly. The measurement was obtained using a Keyence VL-800 stripe-light projection system. After least-squares alignment of the loaded state to the reference state, the center of the bolt head shows a depression of approximately \SI{10}{\micro\meter}--\SI{20}{\micro\meter} relative to the edge region.}
\label{fig:prelim_optical}
\end{figure*}

A third preliminary comparison used a Keyence VL-800 optical 3D measuring device based on stripe light projection. A bolt was measured before and after assembly at \SI{200}{\newton\meter}. The measurement in the assembled state was superimposed on the initial state and fitted using a least-squares alignment. The resulting false-color map shows positive relative displacement in red and negative relative displacement in blue. The center of the bolt head exhibits a depression of approximately \SI{10}{\micro\meter}--\SI{20}{\micro\meter} relative to the edge region, as shown in Figure~\ref{fig:prelim_optical}. The spatial trend is comparable to the FEM result, namely a central depression surrounded by a less-deformed edge region.

These preliminary results support the physical basis of the proposed framework: the bolt-head front face deforms under preload, the deformation is measurable by high-resolution optical 3D measurement, and the strain response is consistent with sensor measurements. At the same time, they define the central risk. A \SI{10}{\micro\meter}--\SI{20}{\micro\meter} shape change is detectable with high-performance metrology, but it is not automatically recoverable from ordinary RGB or smartphone images. The subsequent framework, therefore, treats smartphone-based preload estimation as a feasibility target to be established through controlled image-pair acquisition, surface preparation, registration-quality assessment, and uncertainty-aware prediction.
\section{PreForceVision Framework}

\begin{figure*}[t]
    \centering    \includegraphics[width=0.91\textwidth, trim=0cm 0cm 0cm 0cm, clip, keepaspectratio]{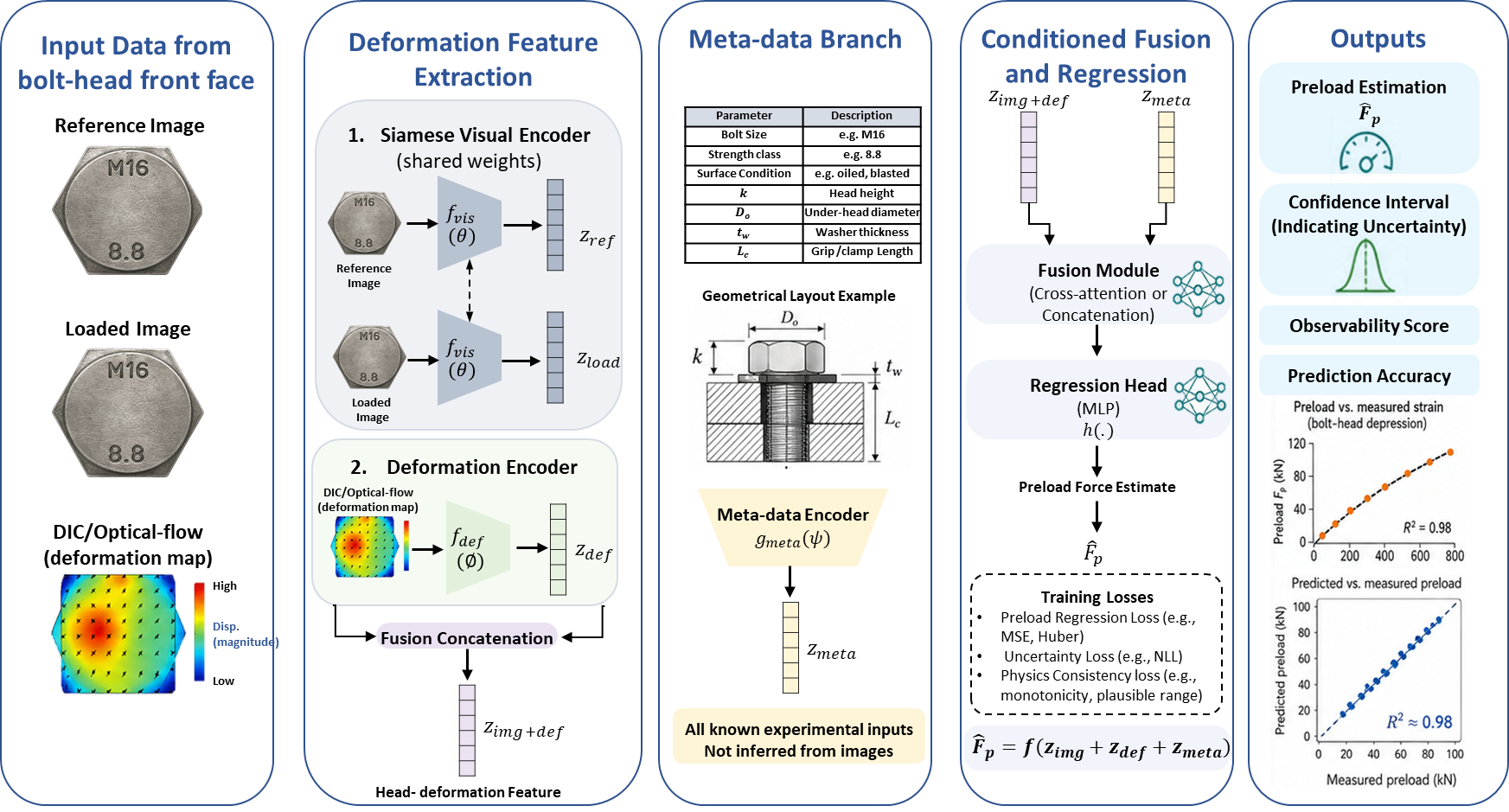}
    \caption{Corrected learning mechanism for preload estimation. Deformation features are extracted from the bolt-head front face. Known geometry or assembly information is used only to condition or calibrate the deformation-to-preload mapping; it is not inferred directly from the head image.}
    \label{fig:model}
\end{figure*}

The framework is organized into five components (Figure~\ref{fig:model}) that move from physics to images and then to prediction. The components are designed to avoid a common pitfall in image-based engineering studies: training a model before proving that the measured image change is mechanically meaningful.
The methodological principle follows an inverse condition-assessment concept: a deformation field measured at an accessible surface is used to infer a non-directly observable mechanical state. In this study, the accessible surface is the bolt-head front face, and the inferred state variable is bolt preload. This requires four steps: selecting a mechanically informative measurement region, generating physics-based and experimental ground truth, extracting repeatable deformation features from reference-loaded image pairs, and mapping these features to preload with uncertainty estimation.

\subsection{Component 1: Mechanical Observability Analysis}
The first component determines whether the bolt-head front face contains a preload-sensitive deformation field. A finite-element model of the selected joint is used to compute the directional displacement normal to the head face, radial and circumferential strain distributions, central-to-edge deformation contrast, and the influence of the controlled assembly configuration. The analysis supports two decisions: the selection of the region of interest and the definition of physically meaningful descriptors.

For the first proof of concept, the geometry is fixed. This is important: the model is not expected to detect washer diameter, head height, or clamping length from the bolt-head image. Those quantities are either controlled by design or supplied as metadata in later multi-geometry studies. The image-derived quantity is the deformation of the head face.

\subsection{Component 2: Multimodal Ground Truth}

Because the optical signal is small, image-based estimates must be validated against independent measurements. The proposed data-generation setup combines: (i) applied preload from the testing machine or load cell; (ii) strain gauges on the bolt-head face or a sensor bolt; (iii) optical 3D measurements of the head surface; (iv) DIC-based displacement or strain fields; and (v) finite-element predictions under matching boundary conditions. These sources do not play identical roles. The load cell supplies the target value, strain gauges verify local surface response, optical 3D measurement verifies out-of-plane deformation, DIC provides full-field optical displacement or strain, and the finite-element model supplies expected spatial trends.

\subsection{Component 3: Controlled Image-Pair Acquisition}

Each specimen is recorded as a reference-loaded image pair $(I_0,I_L)$. In the first study, $I_0$ is recorded before loading or at a defined baseline state, and $I_L$ is recorded after applying a known preload. The reference image is essential because the visible signal is not the absolute appearance of the bolt, but the change of local features on the same surface.

The acquisition should proceed in staged levels of increasing complexity. Stage A uses an industrial camera, fixed stand, fixed lighting, and prepared surface texture to establish the best-case optical measurement limit. Stage B evaluates whether the same deformation information can still be extracted with a simplified low-cost camera setup under controlled laboratory conditions. Stage C introduces controlled perturbations in pose, distance, illumination, surface condition, and camera hardware to quantify robustness. At each stage, metadata such as camera distance, illumination, exposure, and surface preparation are recorded.
\subsection{Component 4: Deformation-Feature Extraction}

\begin{figure}[t]
    \centering
    \includegraphics[width=1.\columnwidth, trim=0cm 0cm 0cm 0cm, clip, keepaspectratio]{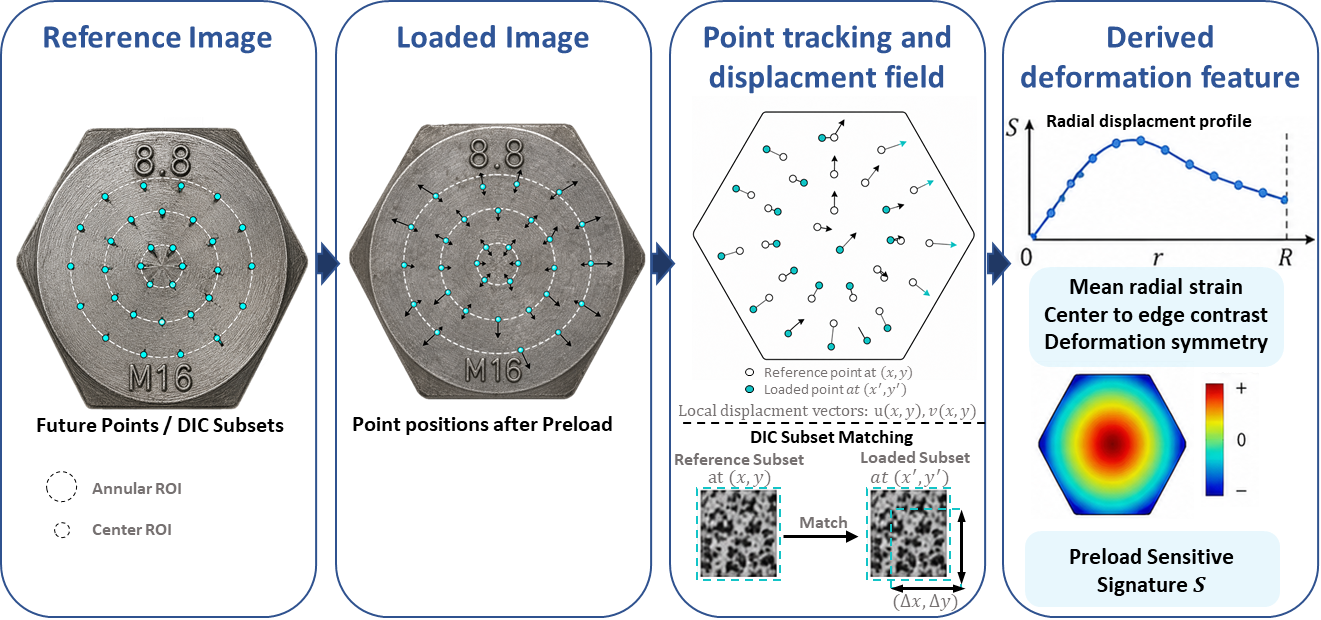}
    \caption{Schematic feature-extraction principle. The model tracks local point motion on the bolt-head front face between the reference and loaded states, then converts this motion into displacement, strain, contrast, and symmetry descriptors.}
    \label{fig:tracking}
\end{figure}

Feature extraction starts with segmentation of the bolt-head front face. The mask $M$ can be obtained from the hexagonal head contour, fiducial marks, or a segmentation model. The reference and loaded images are then registered by correcting rigid translation, rotation, scale, and perspective. This step is critical because rigid motion must not be interpreted as preload-induced deformation. The overall process is shown in Figure~\ref{fig:tracking}.

After registration, photometric normalization reduces illumination differences and specular effects. DIC, optical flow, or feature tracking is then used to estimate the local motion of surface texture or speckle subsets. The local displacement vector is \ref{eq:displacment}
where $(x,y)$ is the subset center in the reference image and $(x+u,y+v)$ is the corresponding position in the loaded image. This produces a displacement field on the bolt-head face.

From the displacement field, physically interpretable descriptors are computed. The most relevant descriptors for this use case are: mean radial displacement $\bar{u}_r$, mean radial strain $\bar{\varepsilon}_r$, center-to-edge displacement contrast $S_{c-e}$, annular strain contrast $S_a$, deformation symmetry $\Gamma$, and DIC correlation quality $Q_c$. A compact feature vector can be written as
\begin{equation}
    \mathbf{s}=\left[\bar{u}_r,\bar{\varepsilon}_r,S_{c-e},S_a,\Gamma,Q_c,Q_I\right]
    \label{eq:feature-vector}
\end{equation}
where $Q_I$ summarizes image-quality indicators such as focus, exposure, reflection area, and registration residual. These quality terms are not preload features; they are used to reject unreliable images or increase prediction uncertainty.

\subsection{Component 5: Physics-Guided Prediction}

The predictive model maps the deformation signature to preload. A suitable architecture has three input streams: an image-pair stream, a deformation-map stream, and a metadata stream. The image-pair stream uses a Siamese encoder with shared weights to process $I_0$ and $I_L$. The deformation stream processes DIC or optical-flow maps. The metadata stream contains known experimental or design variables, such as bolt type, surface preparation, and acquisition conditions. If multiple geometries are included later, geometry is provided here as known information.

The model output is the preload estimate and uncertainty,
\begin{equation}
    \left(\hat{F}_p,\sigma_{\hat{F}}\right)=h_\theta\left(\mathbf{z}_{img},\mathbf{z}_{def},\mathbf{z}_{meta}\right)
\end{equation}
where $\mathbf{z}_{img}$, $\mathbf{z}_{def}$, and $\mathbf{z}_{meta}$ are embeddings from the image, deformation, and metadata branches. For the first study, a simpler and more interpretable baseline should be evaluated first:
\begin{equation}
    \hat{F}_p = g(\mathbf{s})
\end{equation}
where $g$ can be linear regression, Gaussian process regression, random forest, or gradient boosting. The learning-based image model should then be compared against this baseline.

The model must learn three physically motivated rules. First, for fixed geometry and fixed acquisition conditions, the preload estimate should increase monotonically with the magnitude of preload-sensitive deformation descriptors. Second, the estimate should be invariant to rigid image motion and illumination variation after registration and normalization. Third, uncertainty should increase when image quality, registration quality, or DIC correlation quality deteriorates. 

A training loss can combine these requirements:
\begin{equation}
\mathcal{L}=\mathcal{L}_{reg}+\lambda_1\mathcal{L}_{mono}+\lambda_2\mathcal{L}_{phys}+\lambda_3\mathcal{L}_{unc}
\end{equation}
where $\mathcal{L}_{reg}$ penalizes preload error, $\mathcal{L}_{mono}$ penalizes non-physical ordering of predictions within the same specimen, $\mathcal{L}_{phys}$ penalizes disagreement with DIC or finite-element trends, and $\mathcal{L}_{unc}$ calibrates predicted uncertainty.

\section{Experimental Design and Evaluation}
\subsection{Variables}
The framework distinguishes controlled variables, measured variables, and target variables. Controlled variables define the mechanical and acquisition conditions: preload level, bolt specimen, surface condition, camera type, camera pose, illumination, exposure, and acquisition distance. In the preliminary study, joint geometry is fixed. Measured variables include reference images, loaded images, segmentation masks, displacement fields, DIC strain fields, strain-gauge signals, optical 3D displacement maps, and image-quality indicators. Target variables include preload force $F_p$, prediction uncertainty $\sigma_{\hat{F}}$, and an observability score indicating whether the image pair contains sufficient deformation.

\subsection{Proposed Test Matrix}
A practical preliminary study should use a staged test matrix. The mechanical stage applies preload levels such as 0, 30, 60, and \SI{90}{\kilo\newton}, matching the preliminary measurements, and can later be refined with intermediate levels. Each load step is repeated over several loading cycles to quantify repeatability and hysteresis. The optical stage records reference-loaded pairs under fixed camera and lighting conditions. The robustness stage then introduces controlled perturbations: small camera rotations, different illumination angles, surface texture variations, and repeated specimen mounting. Each measurement is linked to a specimen identifier so that image data, preload, strain-gauge data, optical 3D data, and metadata can be traced consistently.

\subsection{Validation Levels}
The validation strategy addresses four questions, \textit{Mechanical validation} asks whether finite-element results, strain gauges, and optical 3D measurements show the same qualitative deformation pattern and a monotonic relation to preload, thereby validating the physical premise; \textit{Optical validation} asks whether DIC or optical-flow features extracted from images correlate with measured preload under controlled acquisition, thereby validating the image measurement chain; \textit{Model validation} asks whether the predictive model estimates preload on unseen specimens or loading cycles, with specimen-wise splitting required to avoid memorizing bolt-specific scratches, stamped markings, or surface artifacts; and \textit{Robustness validation} asks whether predictions remain stable under changes in pose, illumination, camera, and surface condition, where poor image pairs should not only increase error but also increase uncertainty or lower the observability score.


\subsection{Metrics and Ablation Studies}

Accuracy should be reported using mean absolute error (MAE), root mean squared error (RMSE), relative preload error, and coefficient of determination $R^2$. However, these metrics alone are insufficient. The study should also report monotonicity consistency, DIC correlation quality, image registration residual, and uncertainty calibration. For uncertainty, predicted intervals should be checked against empirical coverage.

Ablation studies are essential for scientific credibility. At minimum, the following configurations should be compared: (i) scalar DIC descriptors only; (ii) DIC descriptors plus image-quality metrics; (iii) image-pair Siamese model only; (iv) image-pair model plus DIC map; and (v) image-pair model plus DIC map and metadata. A strong result would show that physics-based descriptors provide a reliable baseline and that learning-based models improve robustness or reduce error without violating physically expected trends.

\section{Discussion}

The proposed PreForceVision framework defines a stepwise route for evaluating whether preload-induced bolt-head deformation can be mechanically observed, optically extracted, and related to preload with sufficient repeatability.

\subsection{Advantages}
The main advantage is mechanical relevance. The target quantity is preload, not only visible looseness or rotation. This links visual inspection to the functional variable governing joint functional behavior. The second advantage is multimodal validation. Finite-element analysis, strain gauges, optical 3D measurement, and DIC provide complementary evidence and reduce the risk of training a model on non-physical image artifacts. The third advantage is interpretability. Deformation descriptors such as radial displacement, radial strain, center-to-edge contrast, and symmetry can be inspected before and after regression. This is important because the deformation signal is small and requires physically traceable evaluation. 

\subsection{Limitations}
The main limitation is the small signal magnitude. The preliminary deformation is in the micrometer range. Although this magnitude is measurable with high-resolution optical 3D systems, recovering it from image pairs requires controlled lighting, stable camera positioning, suitable surface texture, accurate registration, and sufficient image resolution. A second limitation is surface condition. Speckle patterns or matte coatings improve deformation extraction but reduce practical convenience, while untreated metallic heads are more affected by reflection and weak texture. A third limitation is reference dependence. The first-stage method requires a reference image of the same bolt-head surface; absolute preload estimation from a single image would require stronger assumptions and a substantially larger dataset. A fourth limitation is generalization. A model trained on one bolt type, surface condition, and acquisition setup cannot be assumed to generalize without explicit validation.

\subsection{Implications}

The Preliminary study is a mechanically validated optical feasibility study. The recommended sequence is: establish the deformation field using FEM and strain gauges; confirm the central depression using optical 3D measurements; extract DIC or optical-flow features from controlled image pairs; verify monotonicity and repeatability; then train and evaluate regression models. This sequence establishes the scientific basis before evaluating low-cost or field-compatible acquisition.
\section{Conclusion}
This paper presented PreForceVision, a physics-guided framework for estimating bolt preload from deformation of the bolt-head front face. The framework addresses the gap between accurate but hardware-dependent preload sensing and scalable but mechanically indirect vision-based bolt inspection. Instead of treating bolt inspection as an image-classification task, the proposed approach formulates preload estimation as an inverse deformation-to-force problem. Reference-loaded image pairs are used to extract a preload-sensitive deformation signature, which is then mapped to preload under controlled conditions

Preliminary finite-element, strain-gauge sensor integrated bolt, and optical 3D results support the mechanical basis of the concept. They indicate a measurable concave deformation of the bolt-head face under preload, with a central depression in the order of \SIrange{10}{20}{\micro\meter}. The proposed framework, therefore, focuses on measurement discipline: mechanical observability, multimodal ground truth, controlled image-pair acquisition, deformation feature extraction, physics-guided regression, and uncertainty estimation.

Future work will implement the staged experimental plan, quantify the signal-to-noise ratio across surface preparations and camera systems, and compare interpretable DIC-based baselines with learning-based regression models. Particular focus on uncertainty calibration, specimen-wise validation, and robustness under controlled variations in pose, lighting, surface condition, and camera hardware. After these validation steps, the feasibility of low-cost or field-compatible acquisition can be assessed.


\bibliographystyle{asmeconf}
\bibliography{Main/ref}

\end{document}